\documentclass[runningheads]{llncs}
\usepackage[T1]{fontenc}
\usepackage[most]{tcolorbox}
 \usepackage{booktabs}
 \usepackage{multirow}
 \usepackage{subcaption}
 \usepackage{mathtools, nccmath}
 
 \usepackage[dvipsnames,table,xcdraw]{xcolor}
 \usepackage{bbm}
\usepackage{adjustbox}
\usepackage{graphicx,verbatim, amsmath}
\def\eg{\emph{e.g.}}

\def\ie{\emph{i.e.}} 
\usepackage{pifont}
\definecolor{cvprblue}{rgb}{0.21,0.49,0.74}
\usepackage{xspace}
 
\newcommand{\cmark}{\ding{51}}%
\newcommand{\xmark}{\ding{55}}
\newcommand{\yes}{\textcolor{ForestGreen}{\cmark}}
\newcommand{\no}{\textcolor{Bittersweet}{\xmark}}
\begin{document}
\title{Medical Image Understanding Improves Survival Prediction
via Visual Instruction Tuning}
%

\author{Xixi Liu, Jorge Lazo, Andreas Hallqvist, Mikael Johansson, Åse Johnsson, Jonas S Andersson, Ella Äng Eklund, Patrik Sund, Nasser Hosseini, \\Jennifer Alvén, Ida Häggström}
\authorrunning{X. Liu et al.}
%
\institute{Chalmers University of Technology, Gothenburg, Sweden \\
\email{$\{$xixil, idah$\}$@chalmers.se}
 }
  
\maketitle              
\begin{abstract}

Accurate prognostication and risk estimation are essential for guiding clinical decision-making and optimizing patient management. While radiologist-assessed features from CT scans provide valuable indicators of disease severity and outcomes, interpreting such images requires expert knowledge, and translating rich visual information into textual summaries inevitably leads to information loss. 
In this work, we propose a vision–language framework for 3D CT image understanding that leverages large-scale open-sourced CT images paired with radiology reports through visual instruction tuning. This pre-training enables the model to learn clinically meaningful visual–textual representations, which can then be adapted to downstream survival prediction tasks. By incorporating a survival prediction head on top of the pre-trained model, our approach improves survival prediction from CT images and clinical data while generating clinically meaningful language responses to predefined questions. Experimental results demonstrate that our method outperforms baseline methods in survival prediction, particularly, when clinical data alone is less predictive.
The code will be released upon acceptance.


\end{abstract}
\section{Introduction}
\label{sec:intro}

Survival analysis in medicine aims to predict future patient outcomes such as time to disease progression or death, by using patient features from \eg~images and clinical data to model event times and associated risks. Accurate and interpretable survival modelling is crucial to enable risk stratification, guide personalized treatment decisions, and improve prognostic assessment in clinical practice~\cite{nagy2021pancancer,gyHorffy2021survival}.

 Modalities such as radiology images (\eg computed tomography (CT), magnetic resonance imaging (MRI), and positron emission tomograph (PET)), pathology images (\eg~whole-slide images, WSIs), genomics data (\eg~RNA-seq and DNA sequencing), and tabular clinical data (\eg, age, sex, smoking status, and tumor stage) are commonly utilized to perform survival prediction. There has been a series of work focusing on fusing information from genomic data and WSIs~\cite{chen2020pathomic,co_attention,chen2022pan,xu2023multimodal,zhang2024prototypical,CMTA} to improve the performance of survival prediction. 
 However, from a practical perspective, CT scans are routinely performed in clinical workflows for cancer, cardiovascular, and many other diseases. It is uninvasive and readily available, and easier to obtain compared to biopsies and WSIs. Meanwhile, clinical data is abundant in hospital health record systems.  Therefore, in this work, we focus on utilizing multimodal \emph{clinical data with CT images} to improve survival predictions.
 
 Seminal works have explored fusing information from different medical modalities to improve the performance of survival analysis. Such methods include but are not limited to: early fusion~\cite{Anika_multi_data,ZHENG2023109483}, tensor fusion~\cite{chen2020pathomic,orthogonal_fusion}, and attention-guided fusion~\cite{co_attention,CMTA,Kim_LLMguided_MICCAI2024}.  All methods require clinical and/or imaging data with corresponding survival outcomes for supervised training, not to mention handling of the abundant censoring in survival data~\cite{censoring_and_censored}. Meanwhile, most works only predict a single risk score, commonly evaluated by the concordance index~\cite{c_index}(c-index) along with Kaplan-Meier analysis~\cite{Kaplan1992} for visualization. Although a higher c-index indicates better predictive accuracy, clinicians typically find such models uninformative because a scalar risk score alone does not provide an understandable explanation behind its reasoning or treatment decision guidance. Few interpretable methods~\cite{gervelmeyer2024interpretable} aim to provide patch-level visual explanations. In this work, we leverage a large amount of open-sourced CT-image and coupled text report data to first pre-train a 3D CT vision-language model via visual instruction tuning, enabling implicit multimodal representation learning. A survival prediction head is subsequently added to enable accurate risk prediction from CT image and clinical data, while maintaining the ability to generate clinically meaningful responses to pre-defined clinical questions evaluated against report-derived ground-truth answers. 
Our key \emph{contributions} are summarized below:

\begin{itemize}
    \item
We develop a scalable pipeline to construct visual question-answer (VQA) pairs from radiologist-authored CT reports, enabling clinically aligned language supervision.
    \item
We pre-train a 3D CT vision–language model using visual instruction tuning on large-scale CT scans and their radiology reports (\ie, CT-RATE~\cite{ctrate}), learning joint multimodal representation without explicit fusion objectives.

    \item 
We attach a survival prediction head to the instruction-tuned backbone, enabling improved multimodal survival prediction from CT and clinical data, along with structured language responses evaluated against ground-truth answers using BERTScore~\cite{zhang2019bertscore}.

\end{itemize}

 \section{Method}

  \begin{figure*}[t]
    \centering
    \includegraphics[width=1.0\linewidth]{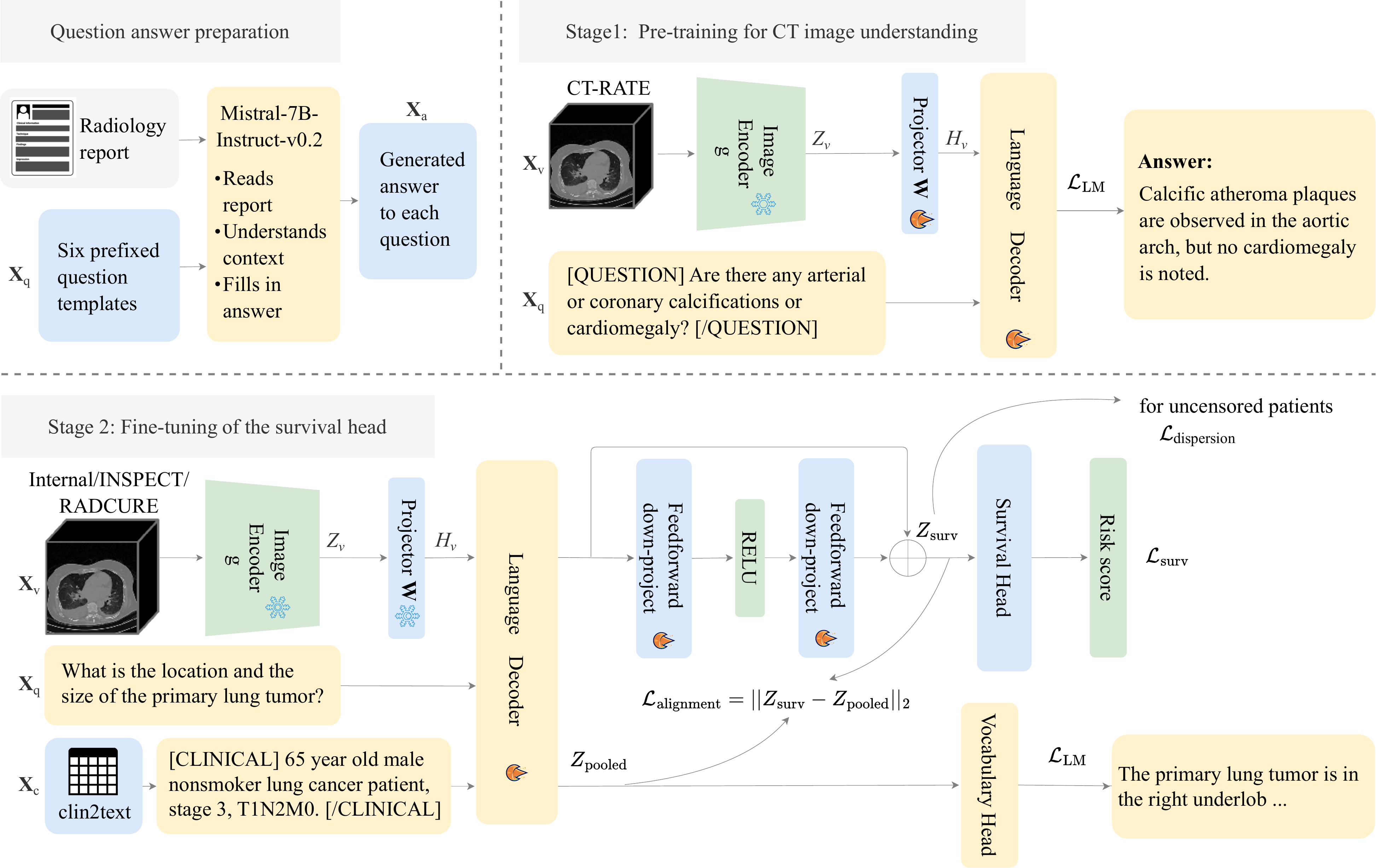}
    \caption{Joint training of visual instruction finetuning and survival head training.}
    \label{fig:placeholder}
\end{figure*}

 \begin{figure*}[h!]
   
\begin{subfigure}[c]{0.48\linewidth} 
   \centering
\includegraphics[width=\linewidth,height=\dimexpr\textheight/4\relax,keepaspectratio]{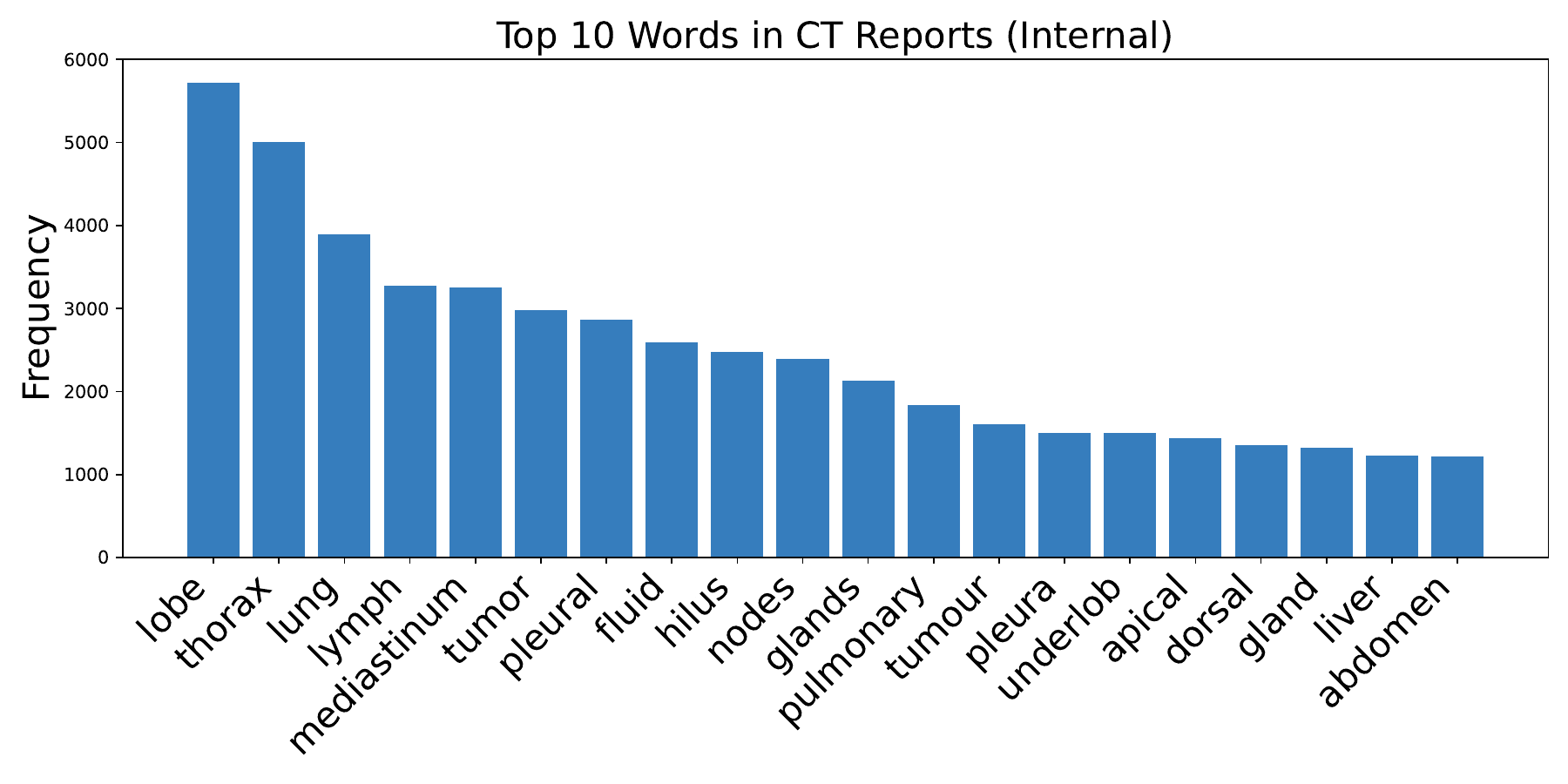}
 \end{subfigure}
  \begin{subfigure}[l]{0.5\linewidth}
 \scriptsize
 \begin{tcolorbox}[title={Predefined radiology questions on the internal lung cancer dataset}, 
                 colback=black!2!white, colframe= cvprblue!65!white,
                  fonttitle=\scriptsize\bfseries ,
   ]
    \begin{enumerate}
        \item What is the location and size of the primary lung tumor?
        \item Which lymph nodes are involved, and what are their size and location?
        \item Is there any pleural involvement or effusion present?
        \item Are metastases or involvement of other organs observed?
        \item Are major vessels involved or compressed?
        \item Are there skeleton/bone changes on the scan?
    \end{enumerate}
\end{tcolorbox} 
 \end{subfigure}
  \caption{Top-10 medical word frequency (left), and the corresponding predefined questions (right).}
   \label{fig:wordfreq}
 \end{figure*}
\subsection{Architecture}
We choose state-of-the-art Radllama~\cite{radllama} as the language decoder $f_\phi(\cdot)$ parameterized by $\phi$, as it has been trained on medical data and has publicly available checkpoints.  For an input 3D CT image $\mathbf{X}_v$, we use the pre-trained Merlin~\cite{blankemeier2024merlin} image encoder denoted by $g$ to extract the visual feature $\mathbf{Z}_v = g(\mathbf{X}_v)$. Specifically, the features are extracted from the last hidden layer of size $7\times7$ in plane, 10 out of plane, and a feature dimension of 2048\footnote{ \scriptsize To reduce computation, the embeddings are averaged over the last (spatial) dimension before projection.}. To connect visual features and the word embedding space, we employ a trainable projection matrix $W$ to transform $\mathbf{Z}_v$ to language embedding tokens $\mathbf{H}_v$, as implemented in~\cite{liu2023llava,llava-med}. In this way, the visual tokens share the same dimensionality as the word embedding space in the language decoder:
$\mathbf{H}_v = \mathbf{W}\cdot \mathbf{Z}_v, \text{with } \mathbf{Z}_v = g(\mathbf{X}_v)$.

Beyond language output, we add a survival branch, denoted $S_\theta(\cdot)$, for survival prediction, which consists of two components: 1) A residual adaptor~\cite{houlsby2019parameter} to balance the quality of the language generation and survival prediction. 2) A specific survival head depending on the survival type (\ie, continuous or discrete).  

\subsection{Construction of question-answer data} \label{data_prep}
 
Given the CT images, corresponding clinical tabular data, and unstructured CT reports, we decompose each report into structured question-answer pairs to facilitate visual instruction tuning. To construct clinically meaningful question templates, we first compute the frequency of medical words across all available CT report and identify the top-100 high-frequency words, these words are then provided to ChatGPT~\cite{openai2024gpt4technicalreport} to generate relevant radiology-style questions templates.  Based on the resulting six predefined templates, we employ Mistral-7B-Instruct-v0.2~\cite{jiang2023mistral7b} to read each report and extract the corresponding answers. To ensure clinical validity and correctness, both the generated templates and the extracted answers were reviewed and approved by two clinicians. For readability, Figure~\ref{fig:wordfreq} presents the top-10 most frequent medical words in our internal lung cancer dataset together with the corresponding predefined question templates. 
\subsection{Training and inference}
 To leverage a large amount of open-sourced 3D CT images paired with radiology report, we employ a two-stage visual instruction-tuning strategy: pre-training stage on CT-RATE dataset~\cite{ctrate} for 3D CT image understanding, followed by fine-tuning for survival prediction with language output on two survival datasets. 
\subsubsection{Stage 1: Pre-training for CT image understanding}
To enhance the proposed framework’s ability to interpret CT images, we construct 282,894 image-text pairs on CT-RATE~\cite{ctrate}, which consists of 17,149 CT scans. For each scan, we perform image grounding by extracting answers to a set of predefined questions from the corresponding radiology report, aligning textual information to relevant image regions. As mentioned before, the image encoder Merlin~\cite{blankemeier2024merlin} and Radllama~\cite{radllama} are both pretrained with medical data, therefore, we skip the training step for medical concept alignment. We directly train the model with curated question-answer pairs extracted from the radiology reports. During training, we keep the visual encoder and survival branch weights frozen, and train only the projection matrix $\mathbf{W}$ and language decoder $f_\phi(\cdot)$. Specifically,  each training sample includes an image $\mathbf{X}_v$, clinical data $\mathbf{X}_c$, a question $\mathbf{X}_q$ related to the CT image with answer $\mathbf{X}_a$. We employ a standard auto-regressive training objective. For an answer with sequence length $L$, the probability of target answers $\mathbf{X}_a$ is
\begin{align}
    p(\mathbf{X}_a| \mathbf{X}_v, \mathbf{X}_\text{instruct}) =\prod_{i=1}^L p_\theta (x_i| \mathbf{X}_v, \mathbf{X}_{\text{instruct}, < i}, \mathbf{X}_{a,<i}),
\end{align}
and the corresponding language training loss reads as 

 \begin{align}
     \mathcal{L}_\text{LM} & = -\log  p(\mathbf{X}_a| \mathbf{X}_v, \mathbf{X}_\text{instruct})    =  -\sum_{i=1}^L \log p_\theta (x_i| \mathbf{X}_v, \mathbf{X}_{\text{instruct}, < i}, \mathbf{X}_{a,<i}),
 \end{align}
where $\theta$ are the trainable parameters, $X_\text{instruct}$ consists of both clinical context, if available, and questions.
\subsubsection{Stage 2: Finetuning the survival head} 
We keep the visual encoder and the pre-trained projection layer frozen, and extend the training to the survival branch, \ie, update the language decoder, residual adaptor, and survival head. Given a dataset $\mathcal{D}=\{X_i, t_i, \delta_i\}$, where $X_i$ denotes the patient-specific information (\eg, clinical or imaging data), and $\delta_i$ represents the corresponding status of the event of interest (\eg, death) observed at time $t_i$. Censorship, $\delta =0$, indicates that the event does not occur during the follow-up observation.  To train the model to understand dataset-specific CT images to make a survival prediction, we curate the paired visual-instruction tuning data as described in Sec~\ref{data_prep} and time-to-event data for joint training. 

Two different well-known survival heads are used: continuous DeepSurv~\cite{deepsurv} and discrete DeepHit~\cite{deep_hit}. For the \emph{continuous} case, the assumption is that patients who experience events earlier are expected to be assigned higher risks. 

 For the \emph{discrete} case, we follow the default DeepHit~\cite{deep_hit} formulation, which models the log-likelihood of the joint distribution of the first event time $k$. Mathematically,

\begin{equation}
\mathcal{L}_{\text{surv}} = \begin{cases}
   -\mfrac{1}{N_{\delta=1}} \sum_{i: \delta_i=1}  (\hat{h}_{\theta_{\text{surv}}}(z_\text{surv}^i)  \nonumber -\log \sum_{j\in\mathcal{R} (T_i)} \exp(\hat{h}_{\theta_{\text{surv}}}(z_\text{surv}^j)), & \text{continuous}\\
    -\mfrac{1}{N} \Bigg[ \delta_i \log (p_{k(i)}(z_\text{surv})) \nonumber  +(1-\delta_i) \log \left ( \sum_{l=c+1}^K p_l(z_\text{surv})\right) \Bigg], & \text{discrete}
\end{cases}
\end{equation}
where $N$ is the number of patients, $\mathcal{R} (T_i)=\{i:T_i\geq t\}$ is the set of patients still at risk of failure at time t, $\delta_i$ represents the corresponding status of event of interest (\eg, death or cancer relapse) observed at time $t_i$, and $K$ is the number of bins for discretizing the event time. For the discrete case, the first loss term accounts for the uncensored patients ($\delta_i=1$) and $k(i)$ is the index for patient $i$ that the event is first observed at $t_{k(i)}$.  It aims to assign a high probability at the true event time. The second term captures the censored patients, and assigns a high probability that the patient survives beyond their censoring time $c$.

 \paragraph{Additional losses}   
 To fully utilize the information from the uncensored patients, a dispersion loss~\cite{cider,wang2025diffusedisperseimagegeneration} is added to encourage the separation among image features of uncensored patients whose survival time differs significantly. For the continuous case, our assumption is that patients with similar survival times are expected to have latent representations that are closer in feature space. For discrete case, patients are partitioned into distinct bins, and we enforce that the mean latent representation of different groups are maximally separated in the embedding space. 

  To encourage this, we define the dispersion loss as

\begin{equation}
\mathcal{L}_{\text{dispersion}} = \begin{cases}
    \mfrac{1}{\sum_{i \neq j} 1} 
\sum_{i \neq j} w_{ij} \, \| z_i - z_j \|_2, & \text{continuous}\\
\mfrac{1}{K} \sum_{i=1}^{K}\log \mfrac{1}{K} \sum_{j=1}^{K} \mathbbm{1}\{j\neq i\} \exp(\mu_i^\top \mu_j/\tau), & \text{discrete}    
\end{cases}
\end{equation}
$z_i$ is the latent representation of patient $i$, $w_{ij} = \exp\left(- \frac{(t_i - t_j)^2}{2 \sigma^2} \right)$ is the weighting term that assigns higher importance to patients with similar survival times, $\tau$ is the temperature parameter, $t_i$ is the observed survival time for patient $i$, corresponding to the time until the event occurs, $\mu_i$ is the mean latent representation of patients belong to group $i$.  Note that we only apply this loss for uncensored cases, \ie, patients with event survival time. Meanwhile, to encourage the language output to be risk-aware, we also add an alignment $\mathcal{L}_2$ loss to regularize the hidden states  $\mathbf{z}_\text{pooled}$, \ie,
\begin{align}
    \mathcal{L}_\text{alignment} = ||z_\text{surv} - z_\text{pooled}||_2,
\end{align}
where $z_\text{pooled}$ is the average hidden state across all tokens. \emph{The final loss} is defined as 

\begin{align}
    \mathcal{L} = \mathcal{L}_{\text{LM}}+ \alpha\cdot \mathcal{L}_\text{surv} + \mathcal{L}_{\text{dispersion}} + \mathcal{L}_\text{alignment}.
\end{align}

$\alpha$ is a hyperparameter to balance the open-ended generation capability and survival prediction. 
 \paragraph{Inference} The same six fixed questions from training are applied to all patients. independent of patient-specific information. The final risk score is the average of the six predictions, forming a simple \emph{ensemble} across prompts to reduce noise and improve robustness.  

\section{Experiments}
We use three datasets CT-RATE~\cite{ctrate}) for pre-training, and INSPECT~\cite{inspect} and Internal for survival head fine-tuning. A summary is provided in Table~\ref{tab:datasets}.

\begin{table}[t]
\centering
\caption{\emph{Summary of datasets,} including anatomical regions, available data types, cohort size, and descriptive statistics for survival analysis.}
\fontsize{8pt}{5pt}\selectfont
\begin{tabular}{lcccccccc}
\toprule
\multicolumn{1}{c}{\bf Dataset} & \multicolumn{1}{c}{\bf Region}   &\multicolumn{4}{c}{\bf  Data modality}  & \multicolumn{1}{c}{\textbf{$\#$ images}} & \textbf{$\#$ censored} \\

 & &   Image &  Report & Clinical & Survival   &   \\ 
\midrule
CT-RATE~\cite{ctrate}  & Chest & \yes  & \yes & \yes & \no  & 47,148   \\
\midrule
Internal & Lung  & \yes   & \yes    &  \yes & \yes  & 9,170 &
950 (22.78\%) \\
INSPECT~\cite{inspect} & Lung  & \yes   & \yes  & \no  &  \yes & 20,078 &
15340 (82.93\%)\\

 \bottomrule
\end{tabular}

\label{tab:datasets}
\end{table}

 \emph{CT images.} We closely follow the pre-processing steps of CT scans as implemented in Merlin~\cite{blankemeier2024merlin}. Specifically, the in-plane axial images are resampled to be 1.5 mm resolution, and the out-of-plane is resampled to be 3 mm. All pixel values were clipped within the Hounsfield unit range of [-1000, 1000]. All input volumes are adjusted to $  224 \times 224 \times 160 $ before processing by the image encoder.
 
 \emph{Clinical data.}  The tabular clinical data is transformed to a sentence, which is regarded as context information by appending it before image tokens. For lung cancer patients (Internal), the covariates utilized are \{`age',  `gender',  `smoking status', `stage', `T stage', `N stage', `M stage'\} \footnote{\scriptsize 64 year-old male former smoker lung cancer patient, stage 1, T stage is T4, N stage is N1, M stage is M0.}. The clinical covariates utilized for pulmonary embolism patients (INSPECT) are \{`age',  `gender',  `pe\_positive',   `pe\_acute',    `pe\_subsegmentalonly', `temperature', `respiratory rate', `mean arterial pressure',`pulse'\}\footnote{ \scriptsize A 61-year-old female, acute pulmonary embolism (not limited to subsegmental arteries), temperature 98.6°F, respiratory rate 18.0 breaths per minute, mean arterial pressure 124.0 mmHg, pulse 65.0 bpm.}.

\paragraph{Implementation details.} For CT slices, we employ the pre-trained Merlin~\cite{blankemeier2024merlin} image encoder to extract image features. For the clinical tabular data that we first convert to text sentences, then utilize the Merlin~\cite{blankemeier2024merlin} text encoder to obtain text features. The language decoder in our framework is Radllama~\cite{radllama}, a medical-adapted version of LLama2-7B~\cite{llama2} fine-tuned on MIMIC~\cite{johnson2016mimic} chest x-ray radiology reports via LoRA~\cite{hu2021lora}. During training, the learning rate is $1e-6$ for models with a language decoder, otherwise $1e-4$. Baseline models use a batch size of 32, while language-decoder models use 8 gradient accumulation steps with a local batch size of 12 (total batch size 96). We use an AdamW optimizer with $\text{betas}=(0.9, 0.999)$, and a cosine learning rate scheduler with 500 warmups, and default $\alpha=0.5$. Experiments are conducted on RTX A6000, using 2 GPUs for training models with a language decoder.

\begin{table}[t!]
\centering
\caption{Survival performance is measured by C-index~\cite{c_index} and language quality by BERTScore~\cite{zhang2019bertscore}. Models without language output are marked as N/A. }
\fontsize{8pt}{5pt}\selectfont
\begin{tabular}{clccccclllll}
\toprule
\bf {Dataset}& \bf Method & \bf Pre-train &\multicolumn{3}{c}{\bf  Modalities} & \bf  c-index & \bf BERTScore  \\
& & &  report &   clinical &   CT-image & &   \\
\midrule
\midrule
\multirow{15}{*}{\rotatebox{90}{\bf Internal}}  & CoxPH~\cite{coxph}        & \no  & \no& \yes   &        & 0.6883  & N/A  \\
&    DeepHit~\cite{deep_hit}    & \no  & \no  & \yes   &      &  0.6884 &  N/A  \\
&      DeepSurv~\cite{deepsurv}    & \no & \no  & \yes   &       & 0.6996 &  N/A  \\
   \cline{2-8} \noalign{\smallskip} 
    
 & DeepHit~\cite{deep_hit}   & \no   & \no &   &   \yes &  0.6329 &  N/A   \\
 &  DeepSurv~\cite{deepsurv}  & \no   & \no &    & \yes &  0.6420 &  N/A \\

  &  Ours (DeepSurv)   & \yes& \yes  &    &  \yes   &  0.6382   &  \underline{0.8656} \\
  &  Ours (DeepHit)   & \yes & \yes  &    &  \yes   &  0.6378    &  \bf{0.8660} \\
 \cline{2-8} \noalign{\smallskip} 
  & DeepHit~\cite{deep_hit}   &\no& \no    & \yes & \yes &  0.6952 &  N/A  \\ 
  & DeepSurv~\cite{deepsurv} &\no & \no  & \yes & \yes & \bf 0.7092  &  N/A    \\

    &  Ours (DeepSurv) &\yes &  \yes & \yes &  \yes & \underline{0.7067}  &  0.8636  \\
   &  Ours (DeepHit)  &\yes& \yes &  \yes & \yes &  0.6852  &0.8629   \\ 
\midrule
\midrule

\multirow{15}{*}{\rotatebox{90}{\bf INSPECT}}  
& CoxPH~\cite{coxph} & \no  & \no& \yes   &          & 0.6474   &  N/A \\
& DeepHit~\cite{deep_hit}    & \no  & \no  & \yes   &        & 0.5567 &  N/A   \\
& DeepSurv~\cite{deepsurv}    & \no & \no  & \yes   &        & 0.5625  &  N/A\\
   \cline{2-8} \noalign{\smallskip} 
    
 & DeepHit~\cite{deep_hit}   & \no   & \no &   &   \yes    &0.7003   &  N/A \\
 &  DeepSurv~\cite{deepsurv}  & \no   & \no &    & \yes   & 0.7030  &  N/A \\

  &  Ours (DeepSurv)   & \yes& \yes  &    &  \yes     &  \underline{0.7145}  & \bf{0.8760} \\
  &  Ours (DeepHit)   & \yes & \yes  &    &  \yes     & 0.7057 & 0.8710 \\
 \cline{2-8} \noalign{\smallskip} 
  & DeepHit~\cite{deep_hit}   &\no& \no    & \yes & \yes   & 0.7008 &  N/A \\ 
  & DeepSurv~\cite{deepsurv} &\no & \no  & \yes & \yes   &  0.7083 &  N/A   \\
     
    &  Ours (DeepSurv) &\yes &  \yes & \yes &  \yes    & \bf 0.7204  & \underline{0.8759}  \\
   &  Ours (DeepHit)  &\yes& \yes &  \yes & \yes &  0.7014  & 0.8746   \\ 
\bottomrule
\end{tabular}
\label{sample-table}
\end{table}
\paragraph{Results and discussions}
We consider three baseline methods, \ie, CoxPH~\cite{coxph}, DeepHit~\cite{deep_hit}, and DeepSurv~\cite{deepsurv}. For DeepHit and DeepSurv, we consider three different settings \ie, text-only, image-only, and both modalities. 

Internal and INSPECT datasets both have CT-images with corresponding radiology reports, therefore they are utilized to demonstrate the effectiveness of visual-instruction tuning for survival prediction. Although not the main task of our model or part of our aim, we here present the performance of the open-ended generation capability via measuring the lexical similarity, \ie, BERTScore~\cite{zhang2019bertscore}. Specifically, we randomly sampled 500 visual question-answer pairs from each dataset, we then compare the answers extracted from the report and the ones generated from our trained models. 

\begin{figure*}[t]
    \includegraphics[width=0.24\linewidth]{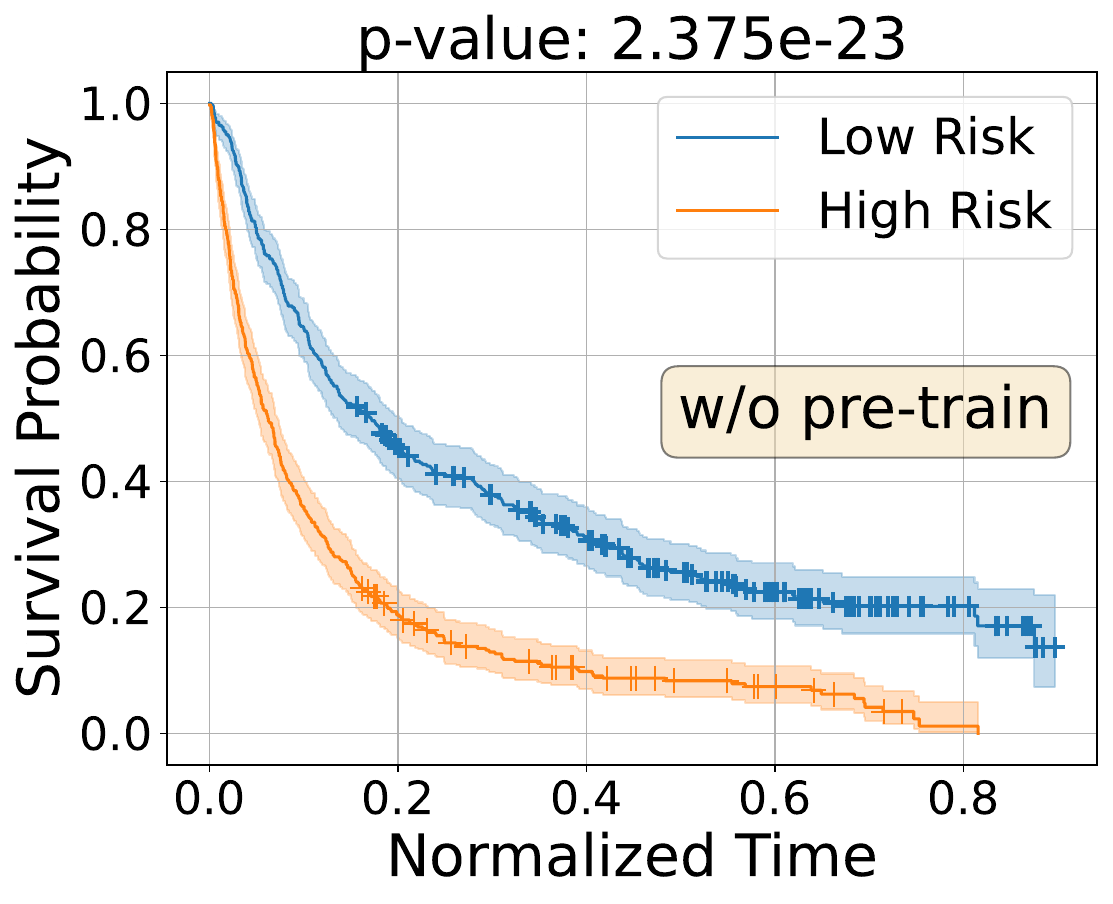}
    \includegraphics[width=0.24\linewidth]{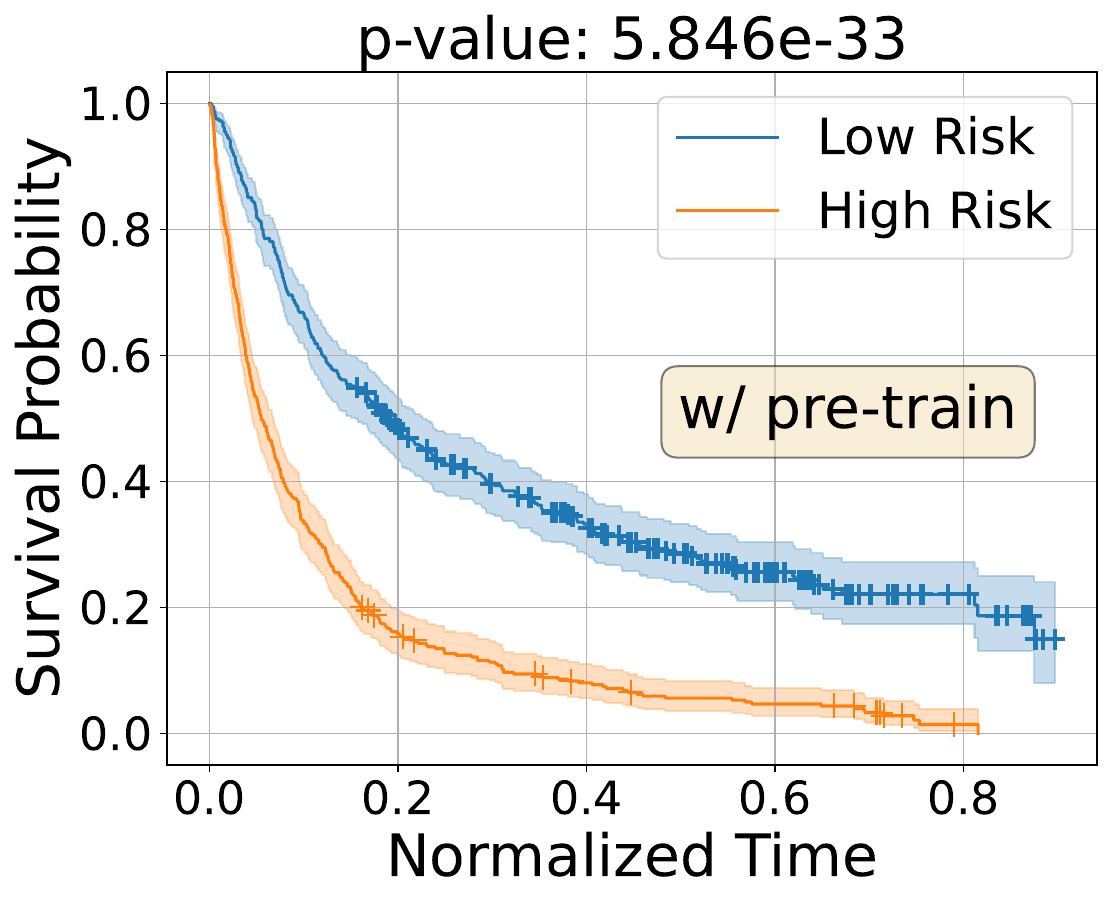}
    \includegraphics[width=0.24\linewidth]{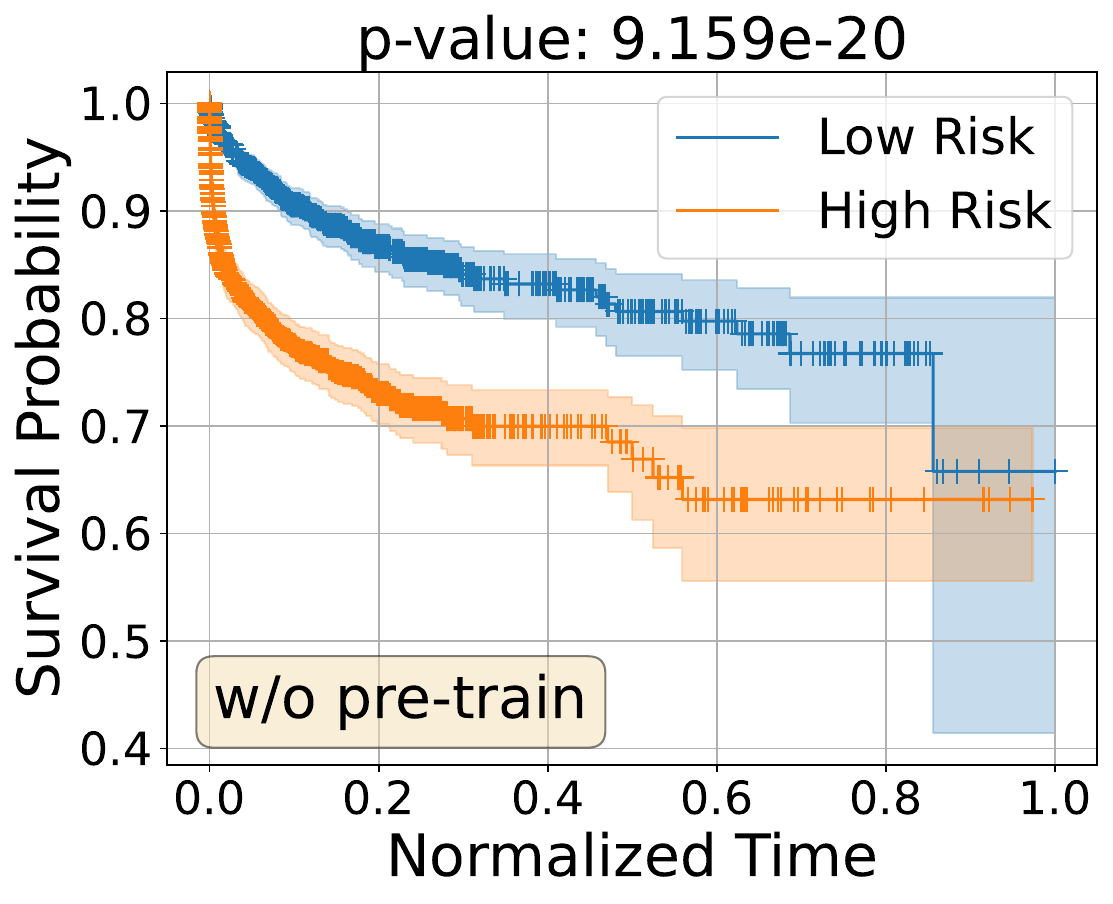}
    \includegraphics[width=0.24\linewidth]{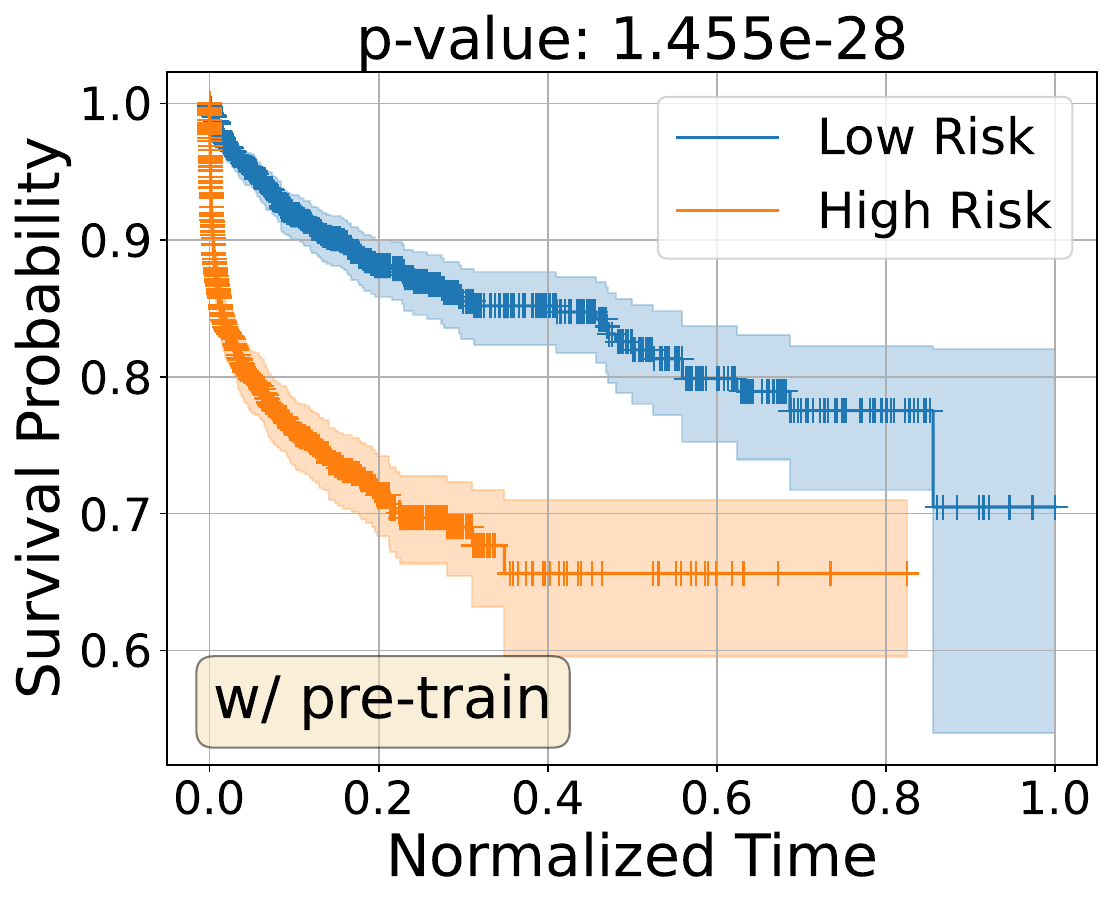}
    
    \caption{Kaplan–Meier plots for Internal (left two columns) and INSPECT (right two columns) datasets, with and without pre-training on CT-RATE~\cite{ctrate} dataset.}
    \label{fig:km}
\end{figure*}
One can see from Table~\ref{sample-table} that our joint training method achieves comparable survival prediction performance with reasonable language outputs on the Internal dataset. A possible reason for the relatively small margin on the Internal dataset is that it focuses on lung cancer, where clinical variables such as TNM stage are highly predictive for survival. In this case, survival performance is already strong when using structured clinical data alone, leaving limited room for improvement from multimodal modeling. In contrast, on the INSPECT dataset, which consists of pulmonary embolism (PE) patients, our method achieves significantly better survival prediction performance, together with strong language generation capability. For PE, CT imaging findings are more critical for risk assessment, while clinical variables such as respiratory rate or mean arterial pressure may be less dominant. Therefore, effectively learning from 3D CT images becomes more important. Our vision-language pre-training and multimodal joint training strategy better exploit imaging information, leading to improved survival prediction. 
In addition, using the predicted risk score of each patient, we divided the cohort into high-risk and low-risk groups, by stratifying at the \emph{median} score. Figure~\ref{fig:km} presents the Kaplan–Meier curves of the Internal and INSPECT test cohorts with and without pre-training on CT-RATE~\cite{ctrate}. Pre-training leads to a clearer separation between the survival curves of the two risk groups, which is further supported by more significant log-rank p-values.

\section{Conclusion}
To our knowledge, we are the first to leverage a large-scale open-sourced 3D CT images paired with radiology reports to pre-train a 3D vision-language model via visual instruction tuning. Our framework improves survival prediction, especially when clinical data is less informative, while generating structured, clinically meaningful language responses.

\subsubsection{Acknowledgment.}
The computations were enabled by resources provided by the National Academic Infrastructure for Supercomputing in Sweden (NAISS), partially funded by the Swedish Research Council through grant 2022-06725. The study was supported by the Sjöberg Foundation grant 2022-489.

\bibliographystyle{splncs04}
\bibliography{references}

\end{document}